\documentclass[12pt]{article}
\usepackage[margin=1in]{geometry}
\usepackage[T1]{fontenc}
\usepackage{lmodern}
\usepackage{graphicx}
\usepackage{booktabs}
\usepackage{array}
\usepackage{tabularx}
\IfFileExists{microtype.sty}{\usepackage{microtype}}{}
\usepackage{caption}
\usepackage{subcaption}
\usepackage{float}
\usepackage{placeins}
\IfFileExists{csquotes.sty}{\usepackage{csquotes}}{}
\usepackage[
    backend=biber,
    style=numeric-comp,
    sorting=none,
    natbib=true,
    giveninits=true,
    maxbibnames=99,
    maxcitenames=2,
    doi=true,
    isbn=false,
    url=false,
    eprint=true
]{biblatex}
\usepackage[hidelinks]{hyperref}

\newcommand{\lex}[3]{\textit{#1 / #2}}
\newcommand{\latinlex}[2]{\textit{#1 / #2}}

\graphicspath{{figures/}}
\DeclareFieldFormat[article,book,inbook,incollection,inproceedings,misc,thesis,unpublished]{title}{#1}
\DeclareFieldFormat[article]{journaltitle}{\mkbibemph{#1}}
\DeclareFieldFormat[inproceedings]{booktitle}{\mkbibemph{#1}}
\AtEveryBibitem{%
    \ifentrytype{inproceedings}{\clearfield{note}}{}%
}

\newbibmacro*{titleauthors}{%
    \printfield[title]{title}%
    \newunit\newblock
    \ifnameundef{author}
        {}
        {\printnames{author}}%
}

\DeclareBibliographyDriver{article}{%
    \usebibmacro{bibindex}%
    \usebibmacro{begentry}%
    \usebibmacro{titleauthors}%
    \newunit\newblock
    \printfield[journaltitle]{journaltitle}%
    \setunit*{\addcomma\space}%
    \printfield{volume}%
    \setunit*{\addcomma\space}%
    \printfield{number}%
    \setunit*{\addcomma\space}%
    \printfield{pages}%
    \setunit*{\addcomma\space}%
    \printfield{year}%
    \newunit\newblock
    \usebibmacro{doi+eprint+url}%
    \usebibmacro{finentry}%
}

\DeclareBibliographyDriver{inproceedings}{%
    \usebibmacro{bibindex}%
    \usebibmacro{begentry}%
    \usebibmacro{titleauthors}%
    \newunit\newblock
    \printfield[booktitle]{booktitle}%
    \setunit*{\addcomma\space}%
    \printlist{publisher}%
    \setunit*{\addcomma\space}%
    \printfield{pages}%
    \setunit*{\addcomma\space}%
    \printfield{year}%
    \newunit\newblock
    \usebibmacro{doi+eprint+url}%
    \usebibmacro{finentry}%
}

\DeclareBibliographyDriver{misc}{%
    \usebibmacro{bibindex}%
    \usebibmacro{begentry}%
    \usebibmacro{titleauthors}%
    \newunit\newblock
    \printfield{howpublished}%
    \setunit*{\addcomma\space}%
    \printfield{note}%
    \setunit*{\addcomma\space}%
    \printfield{year}%
    \newunit\newblock
    \usebibmacro{doi+eprint+url}%
    \usebibmacro{finentry}%
}

\DeclareBibliographyDriver{book}{%
    \usebibmacro{bibindex}%
    \usebibmacro{begentry}%
    \usebibmacro{titleauthors}%
    \newunit\newblock
    \printlist{publisher}%
    \setunit*{\addcomma\space}%
    \printfield{year}%
    \newunit\newblock
    \usebibmacro{doi+eprint+url}%
    \usebibmacro{finentry}%
}

\title{Between Century and Poet: Graph-Based Lexical Semantic Change in Persian Poetry}
\author{
Kourosh Shahnazari\\
\small Sharif University of Technology
\and
Seyed Moein Ayyoubzadeh\\
\small Sharif University of Technology
\and
Mohammadali Keshtparvar\\
\small Amirkabir University of Technology
}
\date{}

\begin{document}
\maketitle

\begin{abstract}
Meaning in Persian poetry is historical, but it is also profoundly relational. Words endure through repeated literary traditions while changing their force through shifting constellations of neighboring terms, rhetorical frames, and poetic voices. This study examines that process by combining aligned Word2Vec spaces with graph-based neighborhood analysis across century slices and across a selected set of major poets. Rather than treating semantic change as vector displacement alone, it models lexical history as the rewiring of local semantic graphs: the loss and gain of neighbors, altered bridge positions, and movement across communities.

A panel of twenty target words structures the analysis, with five recurrent reference words at its center: \lex{khaak}{Earth}{خاک}, \lex{shab}{Night}{شب}, \lex{mey}{Wine}{می}, \lex{baadeh}{Wine}{باده}, and \lex{del}{Heart}{دل}. Around them stand affective, courtly, elemental, and Sufi terms such as \lex{eshgh}{Love}{عشق}, \lex{gham}{Sorrow}{غم}, \lex{darvish}{Dervish}{درویش}, \lex{shah}{King}{شاه}, \lex{fanaa}{Annihilation}{فنا}, and \lex{haqiqat}{Truth}{حقیقت}. Together these words do not participate in a single profile of change. \lex{shab}{Night}{شب} is more time-sensitive than poet-sensitive, \lex{khaak}{Earth}{خاک} is more sharply differentiated by poetic voice, and \lex{del}{Heart}{دل} shows strong continuity despite marked graph-role mobility. The wine pair makes probe choice decisive: \lex{mey}{Wine}{می} is broad, unstable, and semantically noisy, whereas \lex{baadeh}{Wine}{باده} is cleaner, narrower, and easier to interpret.

A lexical audit of affective, elemental, courtly, and Sufi vocabulary confirms that the same corpus contains strongly historical words, strongly poet-shaped words, and sparsely attested mystical terms that demand caution. Across this lexicon, semantic change in Persian poetry is captured more effectively as neighborhood rewiring than as drift in abstract space alone. For Digital Humanities, that shift matters because it restores local structure to computational evidence and allows lexical history to be described in terms closer to literary reading: persistence, migration, mediation, and selective reshaping.
\end{abstract}


\section{Introduction}

Persian poetry is built from lexical recurrence, but recurrence does not imply sameness. A word can return across centuries with unmistakable familiarity and yet do different work each time it appears. The shift may be subtle: a redistribution of neighboring images, a change in emotional register, or a movement from one symbolic field to another. It may also be dramatic, especially when a term travels between epic, mystical, ethical, courtly, and modernizing idioms. Meaning in this tradition is therefore neither purely lexical nor purely contextual. It is relational, sedimented, and historical at once \citep{deFouchecour2006,Schimmel2004Brocade}.

That condition creates a challenge for computational study. Embedding-based models are valuable because they register similarities that remain difficult to capture through manually compiled lexicons alone. Yet embeddings are often read through one dominant question: has a word moved in vector space? Drift is useful, but by itself it offers a thin description of semantic history. A word may remain moderately close to itself while losing one cluster of neighbors, acquiring another, and changing its structural role in the network around it. In poetic language, those local redistributions often matter more than sheer displacement. What changes is not only position, but company.

This matters especially in Persian poetry because literary convention and singular voice are so tightly interwoven. Shared symbolic repertoires endure across long spans of time, but poets do not inhabit those repertoires identically. The same lexical item can become devotional, ethical, cosmological, erotic, or satirical depending on the poet who activates it. Historical period and poetic voice therefore place overlapping pressures on meaning. An account of semantic change that tracks only centuries risks flattening authorial distinctiveness. An account that compares only poets, by contrast, risks losing the deeper historical rhythms within which poetic difference becomes legible \citep{Meisami1987,Lewis2000Rumi}.

The argument developed here begins from that tension between time and poet. It does so by treating words not as isolated points but as nodes in local semantic graphs built from aligned embeddings. The point is not to abandon vector-space comparison. It is to relocate interpretation at the level where poetic semantics becomes more readable: the neighborhood. Which companions persist? Which ties disappear? When does a word remain hub-like, and when does it become a bridge between otherwise separated clusters? When does a semantic field hold together, and when is it rewired from the inside? Those questions are computationally tractable, but they are also literary in substance.

The resulting picture is not one of uniform lexical change. \lex{shab}{Night}{شب} and \lex{baadeh}{Wine}{باده} lean toward historical sensitivity. \lex{khaak}{Earth}{خاک} is more sharply differentiated by poetic voice. \lex{del}{Heart}{دل} remains comparatively stable, yet its continuity is not immobility. The contrast between \lex{mey}{Wine}{می} and \lex{baadeh}{Wine}{باده} is equally revealing: the first is semantically expansive and noisy, the second more disciplined and therefore more interpretable. These outcomes are important in themselves, but they also support a broader methodological claim. Lexical meaning in Persian poetry evolves not only through vector drift but through the rewiring of semantic neighborhoods, and that rewiring is shaped unevenly by both historical time and poetic voice.

The article makes four contributions. First, it offers a unified comparison between century-level and poet-level lexical behavior within one aligned analytical frame. Second, it shows that graph-based measures of neighborhood rewiring, bridge position, and community movement capture change that pure drift obscures. Third, it develops word-specific interpretations for the recurring reference words that organize the study. Fourth, it grounds those interpretations in a twenty-word semantic panel spanning affective, elemental, courtly, and mystical vocabulary, so that the main argument rests on lexical structure rather than on isolated exemplars.

\section{Related Work}

Research on lexical semantic change is still anchored in distributional comparison across time-sliced corpora. Foundational work quantified broad regularities of semantic drift, while later surveys and shared tasks clarified how strongly diachronic claims depend on corpus construction, alignment, and evaluation design \citep{Hamilton2016,Kutuzov2018Survey,Dubossarsky2017,Schlechtweg2020}. Alignment itself is therefore not a technical afterthought but one of the main conditions of interpretation. Comparative studies of embedding-based change detection, together with work on dynamic embeddings and orthogonal mapping, show that the historical reading of a word can shift materially with the reference space and mapping strategy \citep{Shoemark2019RoomToGlo,Bamler2017,Artetxe2018}.

What remains less common is a relational account of change. Graph-based approaches move the focus from self-distance to the fate of local neighborhoods, asking whether a word keeps the same companions, centrality, and cross-community ties even when its aligned vector remains moderately stable \citep{Ma2024GraphClustering}. Contextual models have added finer-grained evidence for sense variation, but they have not removed the need for legible structural comparison, especially in corpora where figurative density is high and stable sense annotation is difficult \citep{Martinc2020Contextual}. For poetry, the crucial question is often not whether a word acquires a wholly new denotation, but whether the symbolic company around it has been recomposed.

In Persian literary computation, the necessary groundwork has emerged from three directions. Corpus-building has made large-scale comparison of classical Persian texts feasible \citep{Raji2024Corpus}. Studies of chronology, clustering, and authorship show that poet-level differentiation can be modeled computationally \citep{RahgozarInkpen2016,RahgozarInkpen2019,Shahnazari2025PARSI}. Recent graph-based work on Persian poetic symbolism further shows that relational measures such as centrality, modular structure, and field formation can recover literary organization that flatter similarity measures miss \citep{Shahnazari2025NAZM,Shahnazari2026DynamicAtlas}. The literary backdrop is equally important: Persian lyric and mystical poetry depend on recurrent symbolic repertories whose force lies in modulation rather than lexical replacement \citep{deFouchecour2006,Schimmel2004Brocade,Chittick2017Rumi}. The gap addressed here is therefore specific. Persian poetry now has the computational and literary scaffolding for large-scale comparison, but it still lacks a lexical semantic account that measures historical pressure and poetic voice within the same relational frame.

\section{Data and Method}

The corpus contains 129{,}451 poems and 1{,}446{,}347 extracted verses distributed across century bins from the third to the fourteenth century. Coverage is uneven. Century~3 contains only 1{,}090 verses and 14{,}319 tokens, while the middle and later centuries are far denser. The temporal analysis therefore retains two views of the material. Full slices preserve the historical record as transmitted in the corpus. Balanced slices downsample the denser centuries to a common target of 628{,}290 tokens, derived from the smallest reasonably populated century slice, so that historical comparison is not driven by sheer volume. The downsampling is deterministic, poet-aware, and round-robin, with seed~42. Sparse slices below the viability threshold, most notably Century~3, are kept in full and treated explicitly as cautions rather than forced into a misleading appearance of comparability.

The poet-level component follows the same logic of selective confidence. A set of fifteen poets ranging from Ferdowsi to Parvin Etesami is used to compare long historical pressures with poet-specific differentiation. Several poets provide robust token support: Ferdowsi contributes 568{,}200 tokens, Sanai 341{,}237, Nezami 316{,}137, Mowlana 896{,}711, Jami 520{,}624, Saeb 1{,}178{,}596, and Bahar 242{,}920. Two culturally central but thinner corpora, Hafez with 76{,}110 tokens and Parvin Etesami with 69{,}742, are retained for descriptive comparison but excluded from the primary poet-side score. The aim is breadth without pretending that all poets are equally modelable.

The corpus remains close to attested poetic surface form. Arabic letter variants are normalized to their Persian counterparts, diacritics and tatweel are removed, punctuation and spacing are regularized, and internal ZWNJ is preserved where it remains lexical. Token boundaries are stripped without stemming, lemmatization, or semantic-family collapsing. Function words and trivial target variants are set aside only when local neighborhoods are compared, so that dense formulaic repetition does not overwhelm the semantic field.

Each century slice and each poet slice is modeled with a separate Skip-gram with Negative Sampling Word2Vec model. The training configuration is fixed at vector size 200, window size 5, minimum count 15, negative sampling 10, 15 epochs, and seed~42, with \texttt{sg=1} and hierarchical softmax disabled. For the two lower-data poet corpora, Hafez and Parvin Etesami, the minimum count is relaxed to 10 so that their models remain usable for descriptive comparison. Alignment is performed with orthogonal Procrustes on a shared anchor vocabulary. The primary temporal analysis uses consecutive alignment, so each century is aligned directly to its immediate predecessor; the first slice in the chain is copied without transformation.

From each aligned model the analysis builds a mutual \(k\)-nearest-neighbor graph with \(k=10\) using cosine similarity. Two words are connected only when each appears in the other's top-10 neighborhood, which suppresses many weak asymmetric ties. Edges are weighted by mean mutual cosine similarity. Communities are detected with weighted greedy modularity, and three graph-role measures are then tracked: degree centrality, community assignment, and bridge score. Bridge score is defined explicitly as the product of two terms: the share of a node's weighted degree that reaches outside its own community and the diversity of external communities touched by those outgoing ties. The measure rises when a word not only reaches outward but does so across multiple semantic zones.

Century-level change is summarized through adjacent self-drift, neighbor turnover, and graph-role volatility. Drift captures movement between adjacent aligned slices. Neighbor turnover is computed from top-10 neighborhood overlap, so that high scores indicate local rewiring even when cosine self-distance remains moderate. Graph-role volatility combines community movement with normalized changes in degree centrality and bridge score. Degree centrality and bridge score are then read in two complementary ways: absolute century values register field-wide rises and falls, while within-century deviations show which words stand above or below the prevailing level of a given period. A second temporal baseline comes from a single global reference Word2Vec model trained on the concatenated full-corpus slices with the same SGNS architecture, vector dimensionality, window, and lexical threshold used for the century models. Each raw full-century model is aligned directly to that reference by orthogonal Procrustes on a shared anchor vocabulary, and reference deviation is then measured as one minus cosine similarity between the aligned century vector and the reference vector for the same word. Adjacent drift therefore asks whether a word jumps from one century to the next, while reference deviation asks whether the same century also departs from the broader lexical norm of the corpus. When the local jump is high but the reference deviation remains modest, the signal is treated more cautiously as a slice-bound fluctuation, especially near the sparsest early centuries. This comparison preserves historical specificity while tempering the tendency of thin slices to overstate abrupt change. For the agreement profile reported later, high local drift and high reference deviation are defined relative to the median full-panel value of each measure. Poet-level differentiation is measured in parallel through pairwise cosine dispersion and pairwise neighborhood-overlap dispersion across the poet set. Because some poets are broadly similar to many others, the poet-to-poet cosine matrix is also double-centered: each cell is adjusted by subtracting the mean pairwise similarity of its row poet and column poet, then adding back the grand mean. The transformation brings relative affinity into clearer view while leaving the raw cosine values available in the summary tables and poet-side scores. These components are then joined in a century-versus-poet comparison layer that classifies each probe as more time-sensitive, more poet-sensitive, or mixed. A simple cosine-drift baseline is retained throughout as a comparator: it remains useful for movement in embedding space, but it misses cases in which a word's local relations are being rebuilt more sharply than its self-similarity would suggest.

The lexical panel was chosen for interpretability as much as for cultural salience. \lex{khaak}{Earth}{خاک}, \lex{shab}{Night}{شب}, \lex{del}{Heart}{دل}, and the wine pair \lex{mey}{Wine}{می} and \lex{baadeh}{Wine}{باده} all carry long literary histories, but they do not behave identically. That divergence is precisely the point. The study does not seek a single representative word for Persian poetic semantics. It seeks a set of targets capable of revealing different mixtures of persistence, rewiring, and authorial inflection. The panel therefore includes \lex{eshgh}{Love}{عشق}, \lex{jaan}{Soul}{جان}, \lex{cheshm}{Eye}{چشم}, \lex{aatash}{Fire}{آتش}, \lex{darvish}{Dervish}{درویش}, \lex{shah}{King}{شاه}, \lex{rah}{Path}{راه}, \lex{saba}{Breeze}{صبا}, \lex{gham}{Sorrow}{غم}, \lex{rooz}{Day}{روز}, \lex{fanaa}{Annihilation}{فنا}, \lex{baqaa}{Subsistence}{بقا}, \lex{tariqat}{Path}{طریقت}, \lex{haqiqat}{Truth}{حقیقت}, and \lex{soofi}{Sufi}{صوفی} alongside the recurrent reference words, and all of them are traced through the same figures, rankings, and tables.

Robustness is treated here as careful reporting rather than as a large auxiliary experiment. The main temporal interpretation was checked against the balanced slices, against reference-chained alignment in addition to the primary consecutive alignment, and against alternative deterministic resampling seeds. Qualitative interpretation remained stable across those checks. Sensitivity to \(k\) and embedding size was also inspected at a modest level: the article reports the default \(k=10\) graph and 200-dimensional embeddings because they produce the clearest stable comparison, but the central readings do not depend on one arbitrary threshold.

\section{Results}

\subsection{Patterns of Semantic Change}

\begin{figure}[!htbp]
    \centering
    \includegraphics[width=0.92\textwidth]{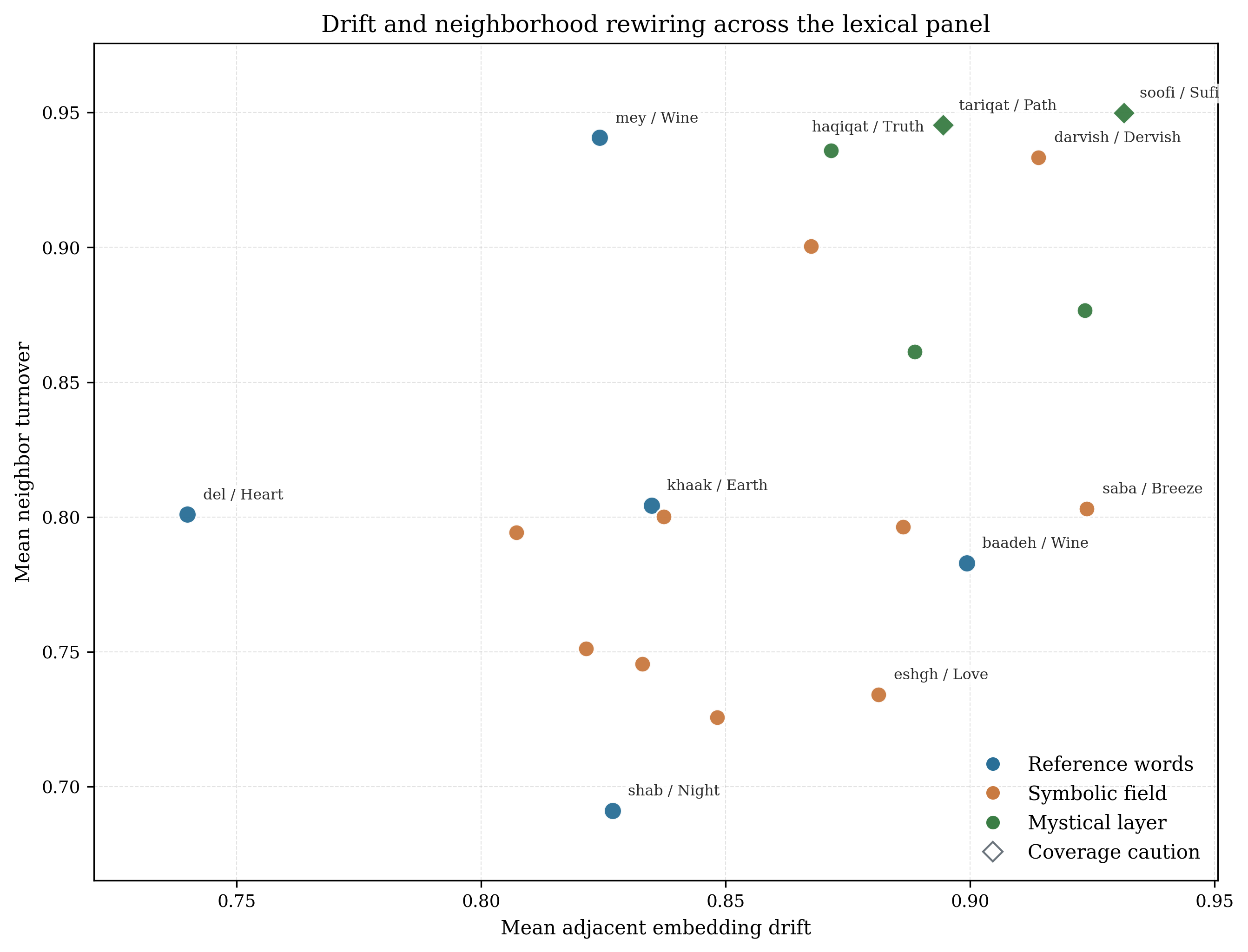}
    \caption{Mean embedding drift against mean neighbor turnover for all twenty target words. Colors distinguish the recurrent reference words, the broader symbolic field, and the explicitly mystical layer; open diamonds mark words whose poet-side coverage remains thin.}
    \label{fig:drift-turnover}
\end{figure}

The twenty-word panel does not sort into a simple opposition between permanence and change. Figure~\ref{fig:drift-turnover} shows that almost every target word accumulates more neighbor turnover than raw self-drift alone would predict. In Persian poetry, lexical history therefore appears less as abrupt displacement than as repeated local reattachment. Words remain recognizably themselves while the company that gives them symbolic force is redistributed. That is precisely what one would expect in a tradition whose major images persist across centuries yet are continually reweighted by genre, doctrine, and voice.

The most volatile edge of the plane is occupied not only by \lex{mey}{Wine}{می} but also by \lex{darvish}{Dervish}{درویش}, \lex{haqiqat}{Truth}{حقیقت}, \lex{tariqat}{Path}{طریقت}, and \lex{soofi}{Sufi}{صوفی}. That cluster is revealing. Broad wine vocabulary and explicitly technical Sufi vocabulary are both prone to strong neighborhood replacement, but for different reasons. \lex{mey}{Wine}{می} diffuses across convivial, mystical, satirical, and lyrical registers; \lex{haqiqat}{Truth}{حقیقت} and \lex{tariqat}{Path}{طریقت}, by contrast, become most active where doctrinal discourse is dense and least evenly distributed. At the more compact end stand \lex{eshgh}{Love}{عشق}, \lex{gham}{Sorrow}{غم}, \lex{rah}{Path}{راه}, \lex{baadeh}{Wine}{باده}, and \lex{del}{Heart}{دل}. These words are not motionless. Rather, their reattachments remain more constrained and therefore more narratable within a stable literary horizon.

\begin{figure}[!tbp]
    \centering
    \includegraphics[width=0.86\textwidth]{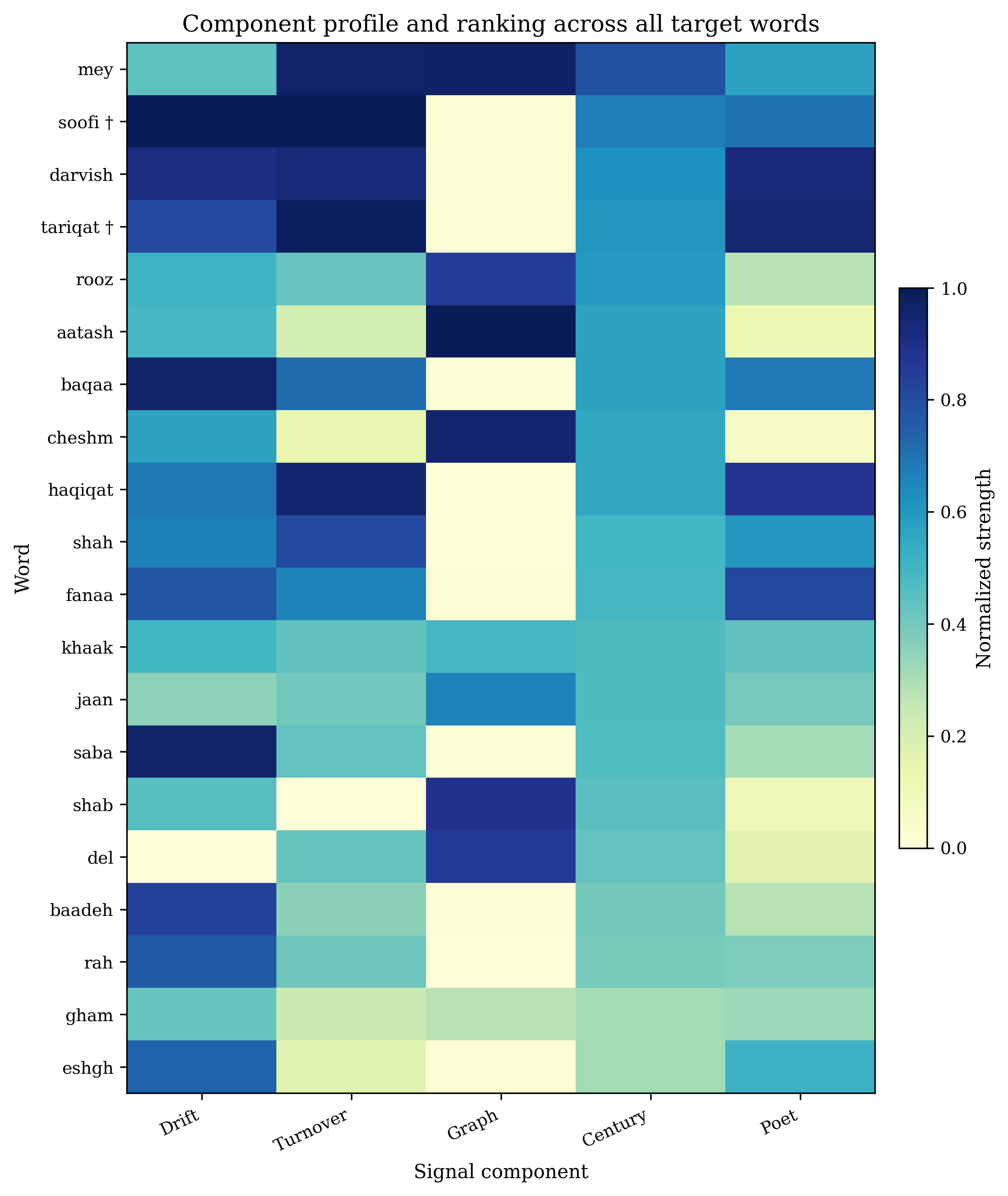}
    \caption{Normalized component profile for the full word set, ordered by century-side signal. The columns report drift, turnover, graph-role volatility, century signal, and poet signal; a dagger marks thin poet-side coverage.}
    \label{fig:signal-profiles}
\end{figure}

Table~\ref{tab:signal-summary} makes the ranking explicit. The most stable relational profiles belong to \lex{eshgh}{Love}{عشق}, \lex{gham}{Sorrow}{غم}, \lex{rah}{Path}{راه}, \lex{baadeh}{Wine}{باده}, and \lex{del}{Heart}{دل}; the strongest century-side reshaping appears in \lex{mey}{Wine}{می}, \lex{soofi}{Sufi}{صوفی}, \lex{darvish}{Dervish}{درویش}, \lex{tariqat}{Path}{طریقت}, and \lex{rooz}{Day}{روز}. This ranking separates emotional durability from lexical stillness. \lex{eshgh}{Love}{عشق} and \lex{gham}{Sorrow}{غم} preserve strong affective centers even when their companions shift. \lex{rooz}{Day}{روز}, by contrast, is historically labile because temporal vocabulary is continually redrawn against \lex{shab}{Night}{شب}, dawn, waiting, judgment, and separation.

\begin{table}[!htbp]
    \centering
    \small
    \caption{Ranking summary for the full target set. The left block lists the most compact relational profiles, the middle block the strongest century-side reshaping, and the right block the words whose poet imprint most clearly outweighs the century signal.}
    \label{tab:signal-summary}
    \begin{tabularx}{\textwidth}{@{}l c l c l c@{}}
\toprule
\multicolumn{2}{@{}l}{Most stable relational profiles} & \multicolumn{2}{l}{Strongest historical reshaping} & \multicolumn{2}{l@{}}{Strongest poet imprint} \\
\cmidrule(r){1-2}\cmidrule(lr){3-4}\cmidrule(l){5-6}
Word & Score & Word & Score & Word & Ratio \\
\midrule
\latinlex{eshgh}{Love} & 0.31 & \latinlex{mey}{Wine} & 0.79 & \latinlex{fanaa}{Annihilation} & 0.59 \\
\latinlex{gham}{Sorrow} & 0.31 & \latinlex{soofi}{Sufi} & 0.67 & \latinlex{eshgh}{Love} & 0.60 \\
\latinlex{rah}{Path} & 0.39 & \latinlex{darvish}{Dervish} & 0.62 & \latinlex{haqiqat}{Truth} & 0.62 \\
\latinlex{baadeh}{Wine} & 0.40 & \latinlex{tariqat}{Path} & 0.60 & \latinlex{tariqat}{Path} & 0.64 \\
\latinlex{del}{Heart} & 0.43 & \latinlex{rooz}{Day} & 0.59 & \latinlex{darvish}{Dervish} & 0.67 \\
\bottomrule
\end{tabularx}

\end{table}

\subsection{Graph-Based Dynamics}

The graph measures clarify how this reattachment proceeds because they separate collective temporal movement from relative lexical differentiation. Figure~\ref{fig:role-trajectories} places absolute century values beside within-century deviations, making it possible to distinguish broad reorganization of the lexical field from the relative prominence of particular words. The shared rises and falls are not noise to be discounted. When many words become more central or more bridge-like together, the semantic field itself is being reorganized. The centered view then shows which words stand above or below that shared baseline. Three broad tendencies appear. First, affective and nocturnal words such as \lex{shab}{Night}{شب}, \lex{del}{Heart}{دل}, \lex{jaan}{Soul}{جان}, and \lex{cheshm}{Eye}{چشم} participate in broad middle- and late-century increases in connectivity, which is consistent with the growing density of lyric interiority, waiting, sight, and separation imagery; yet they do not dominate equally in every period. Second, elemental and itinerant words such as \lex{khaak}{Earth}{خاک}, \lex{rah}{Path}{راه}, \lex{saba}{Breeze}{صبا}, and \lex{rooz}{Day}{روز} move with the wider field in raw terms but oscillate more sharply around the century mean, suggesting repeated redistribution across didactic, cosmological, and symbolic tasks rather than a single historical ascent or decline. Third, the mystical layer becomes structurally active above all from Century~9 onward: bridge scores rise when explicitly Sufi discourse thickens, and \lex{haqiqat}{Truth}{حقیقت} and \lex{tariqat}{Path}{طریقت} become especially distinctive as connectors among doctrinal clusters.

\begin{figure}[!tbp]
    \centering
    \begin{subfigure}[t]{\textwidth}
        \centering
        \includegraphics[width=0.98\textwidth,height=0.77\textheight,keepaspectratio]{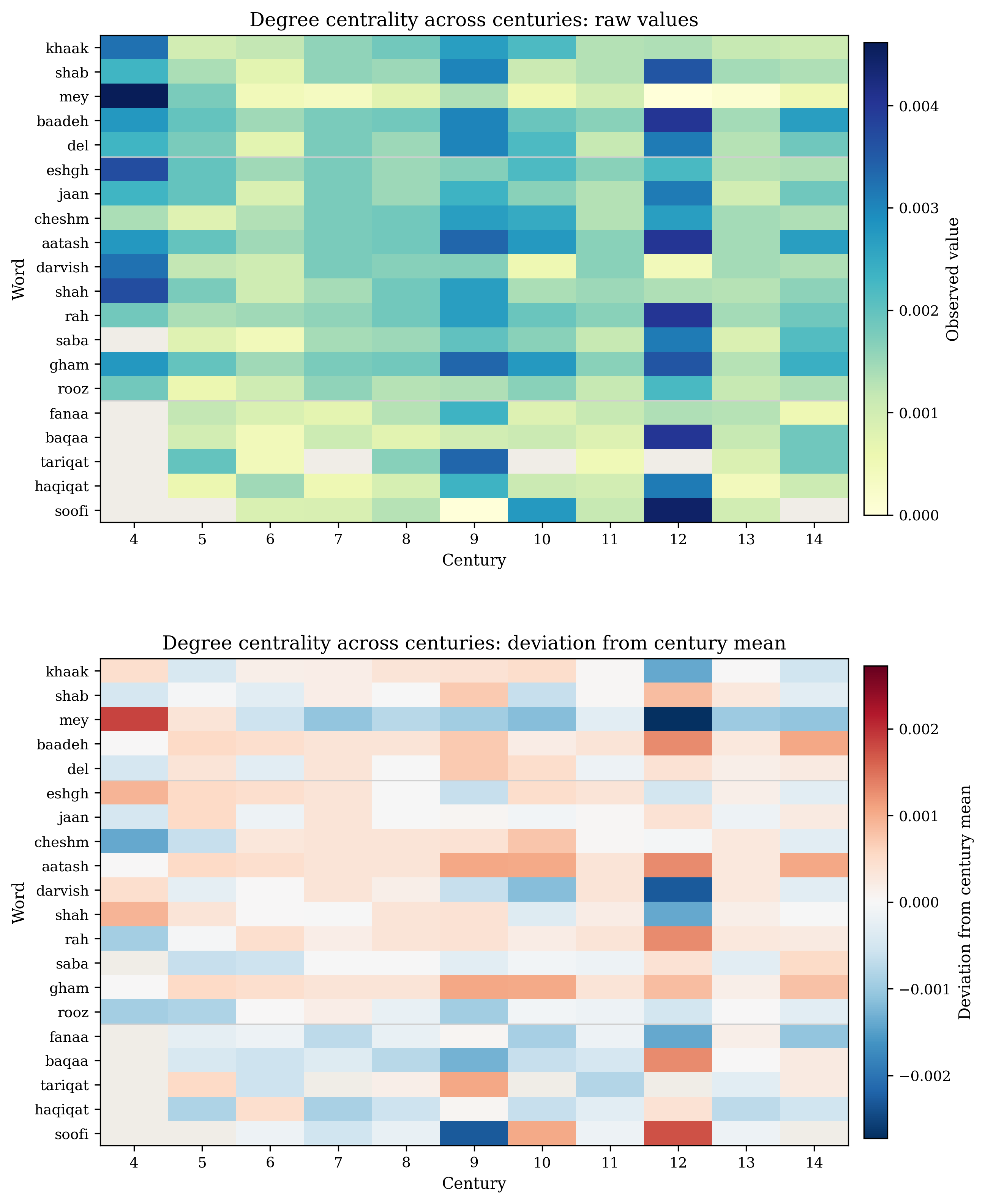}
        \caption{Degree centrality, with raw values above and century-centered deviations below.}
    \end{subfigure}
    \caption{Century-wise graph-role trajectories for all twenty target words. The upper heatmap retains absolute century values, while the lower heatmap centers each century on its mean so positive and negative values mark relative prominence within that period. Century~3 is omitted because its exceptionally thin graph does not sustain comparison with the denser periods that follow.}
    \label{fig:role-trajectories}
\end{figure}

\begin{figure}[!tbp]
    \ContinuedFloat
    \centering
    \begin{subfigure}[t]{\textwidth}
        \centering
        \includegraphics[width=0.98\textwidth,height=0.77\textheight,keepaspectratio]{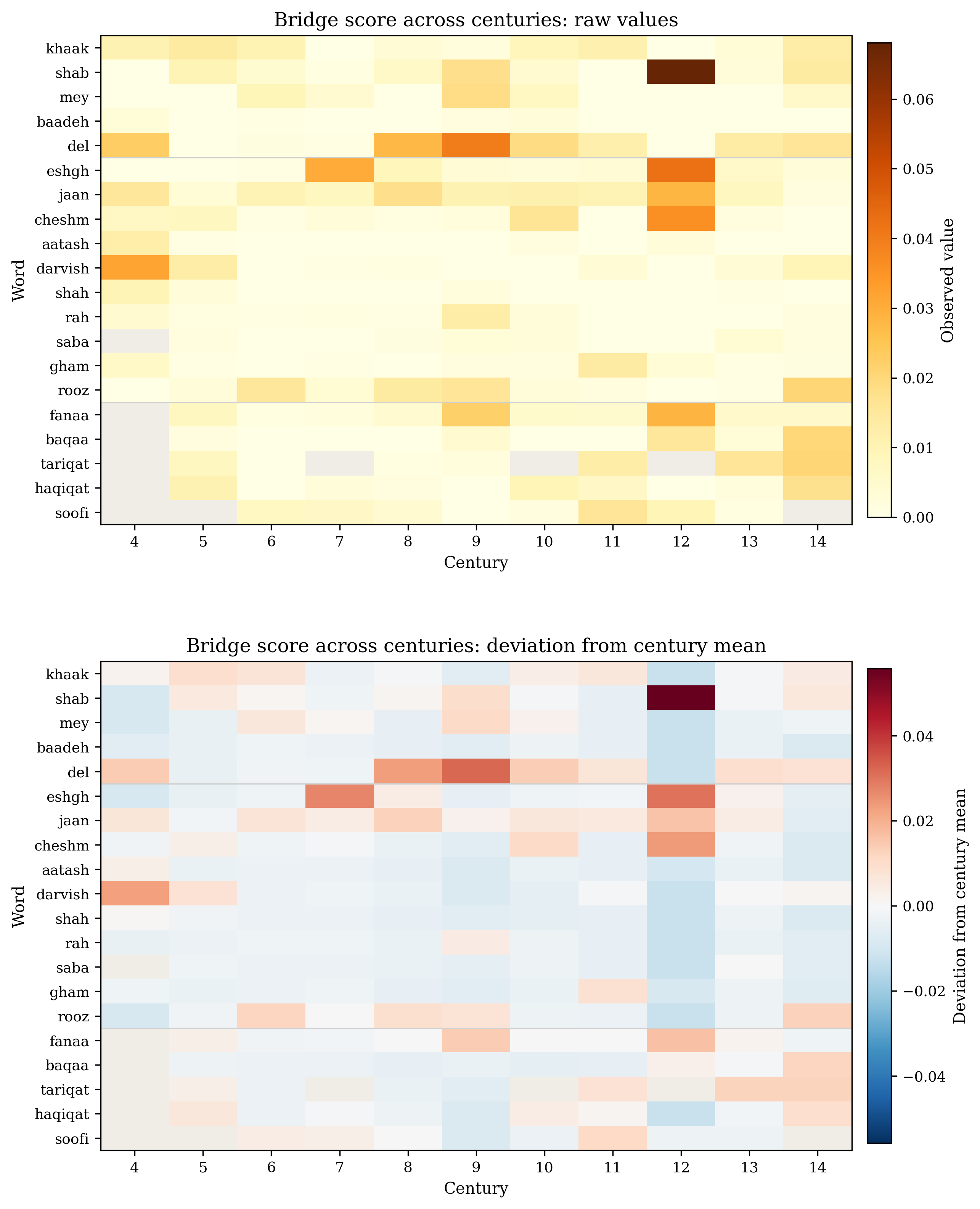}
        \caption{Bridge score, with raw values above and century-centered deviations below.}
    \end{subfigure}
    \caption[]{Century-wise graph-role trajectories for the full word set, continued.}
\end{figure}

\begin{figure}[!tbp]
    \centering
    \includegraphics[width=\textwidth]{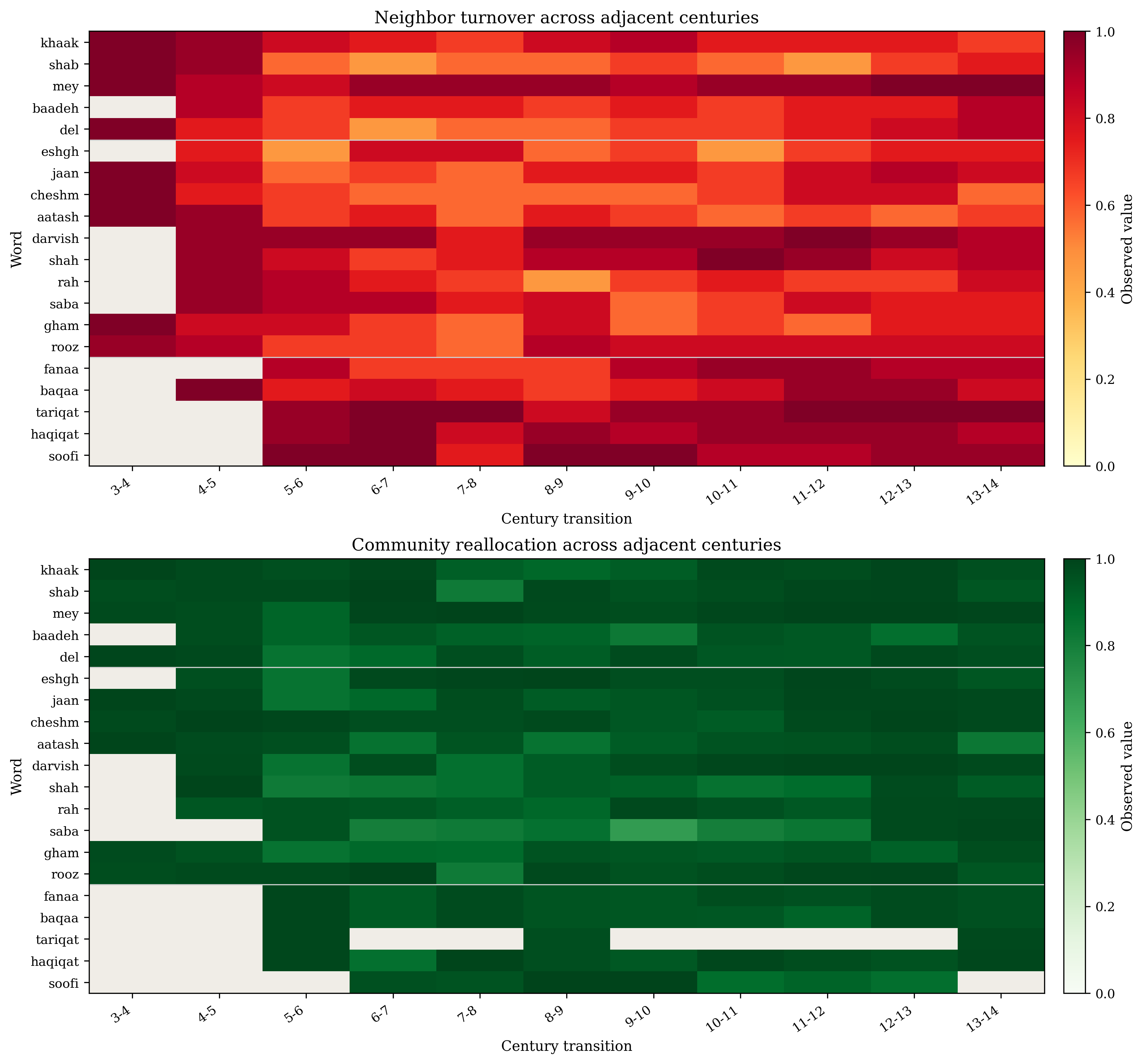}
    \caption{Transition-level dynamics for the full target set. The upper panel reports neighbor turnover between adjacent centuries; the lower panel reports the extent of community reallocation, measured as one minus the overlap between adjacent communities.}
    \label{fig:transition-trajectories}
\end{figure}

Figure~\ref{fig:transition-trajectories} shows why these graph roles matter. Drift alone could suggest a single moving target, but turnover and community reallocation separate words that keep a durable semantic field from words that cross into new coalitions altogether. Neighbor turnover remains high almost everywhere, yet the intensity of community reallocation varies considerably. \lex{mey}{Wine}{می}, \lex{darvish}{Dervish}{درویش}, \lex{shah}{King}{شاه}, \lex{haqiqat}{Truth}{حقیقت}, and \lex{tariqat}{Path}{طریقت} do not simply exchange near-synonyms; they move across markedly different local coalitions because each is pulled among competing social or doctrinal registers. The affective cluster behaves otherwise. \lex{del}{Heart}{دل}, \lex{jaan}{Soul}{جان}, \lex{gham}{Sorrow}{غم}, and even \lex{eshgh}{Love}{عشق} repeatedly rebuild their neighborhoods, but they do so within a recognizable field of desire, grief, self-offering, wound, and inward speech. The words therefore remain legible while their internal weights change, which is precisely the kind of continuity that raw drift alone would flatten.

\begin{figure}[!tbp]
    \centering
    \includegraphics[width=0.98\textwidth]{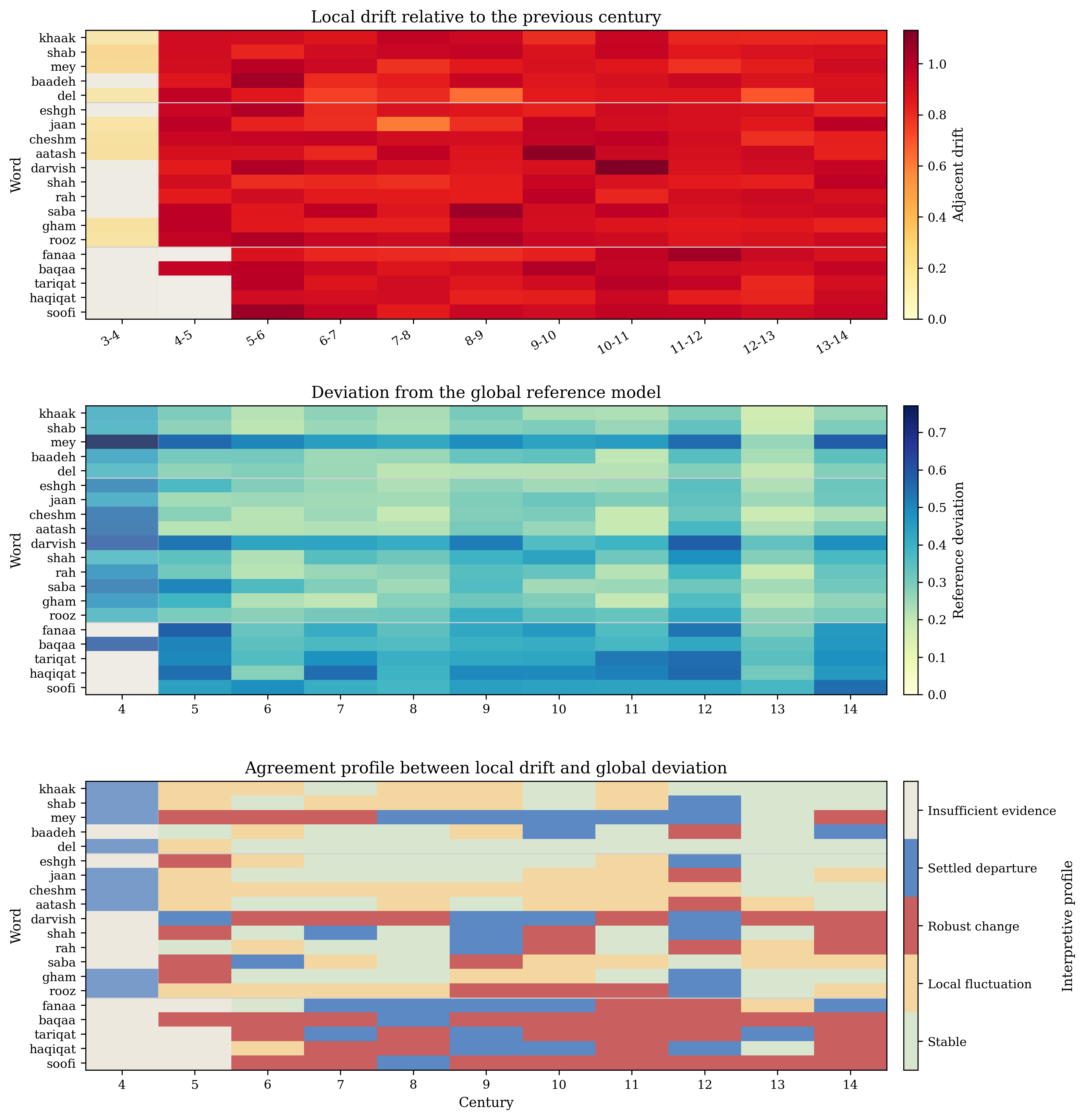}
    \caption{Local century-to-century drift, deviation from the full-corpus reference model, and their agreement profile for all twenty target words. Adjacent drift is assigned to its resulting century. The lower panel classifies each word-period pairing by the joint behavior of the two measures, distinguishing stable usage, local fluctuation, robust change, and settled departure.}
    \label{fig:local-global}
\end{figure}

Sparse centuries require an additional temporal check. Figure~\ref{fig:local-global} compares local adjacent drift with deviation from the global reference model. The distinction matters because an abrupt transition can arise either from durable lexical departure or from the instability of a thin slice. Where both measures are high, the change survives comparison with the full-corpus baseline. Where local drift spikes but reference deviation remains comparatively modest, the movement is better read as local fluctuation. The earliest transitions are the clearest instances of the latter pattern, which is why Century~3 remains visible but analytically constrained rather than allowed to dominate the historical narrative.

The paired comparison tempers several sharp early jumps without flattening the corpus into uniform caution. Broadly shared words can move abruptly at the boundary of a sparse slice and yet remain relatively close to the global reference, whereas lexically specialized items whose local drift is matched by strong reference deviation look more like robust historical departure. The global baseline therefore improves interpretability without replacing the full and balanced century analyses: it shows when an apparent jump is anchored in the wider corpus and when it is better treated as a local perturbation.

\subsection{Case Studies}

\begin{figure}[!tbp]
    \centering
    \includegraphics[width=0.96\textwidth]{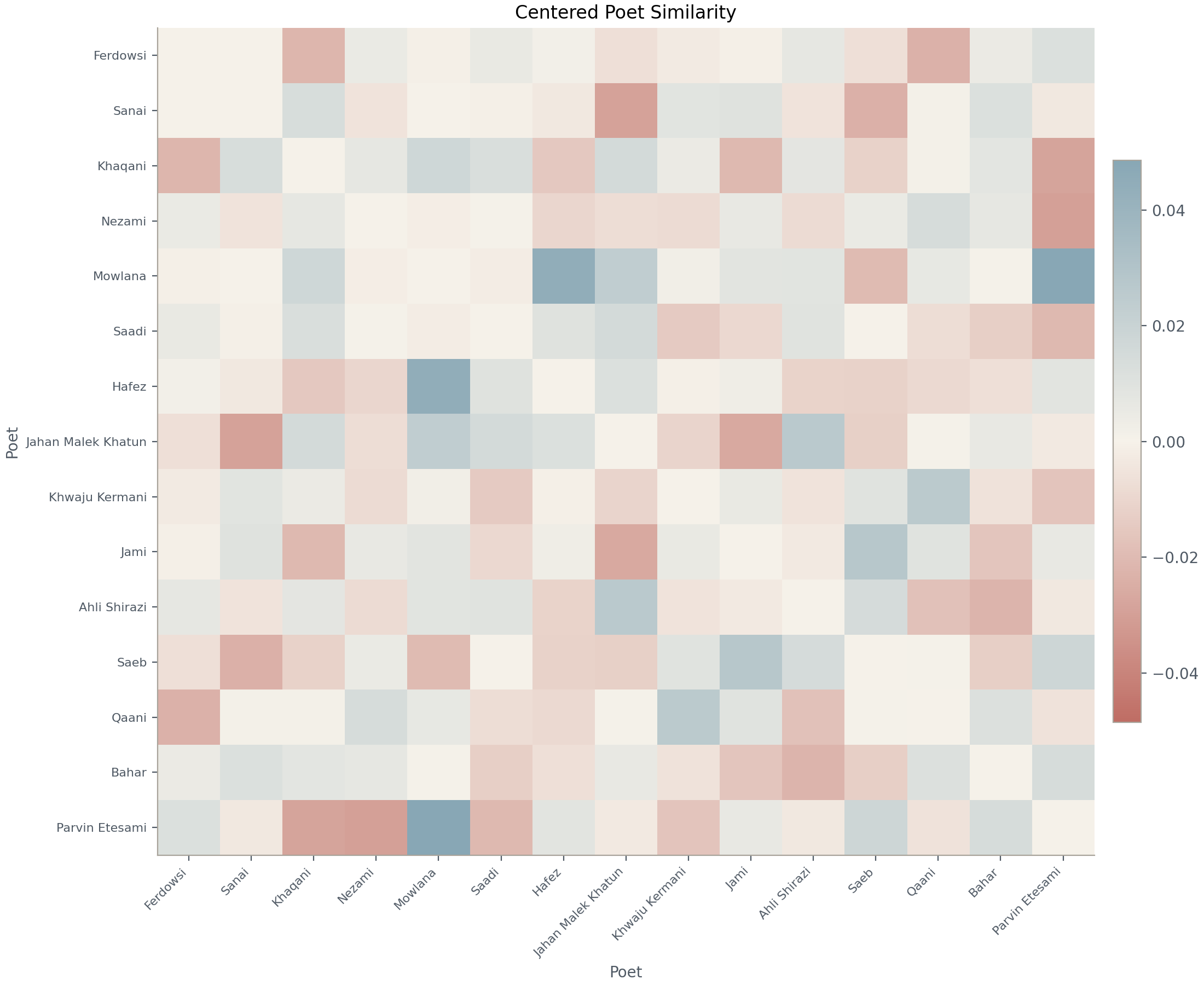}
    \caption{Overall similarity structure across the poet set, shown as a centered version of the raw mean-cosine matrix. Positive values indicate poet pairs whose affinity exceeds each poet's general similarity to the wider set; negative values indicate comparatively weaker affinity.}
    \label{fig:poet-heatmaps}
\end{figure}

The poet-side evidence confirms that different lexical domains respond differently to authorship. Figure~\ref{fig:poet-heatmaps} gives the centered similarity structure of the poet set as a whole, while Table~\ref{tab:focus-comparison} condenses the five recurrent reference words into one comparative profile of drift, rewiring, raw hub level, and century-centered hub deviation. Read together, they show not merely that the reference words differ, but how they differ: some remain broadly shared and historically recurrent, whereas others draw much more of their force from poet-specific redistribution. \lex{khaak}{Earth}{خاک} remains the clearest poet-shaped reference case, and the same logic extends across the rest of the lexicon. \lex{eshgh}{Love}{عشق} moves between reproach, lyric longing, and mystical attraction; \lex{darvish}{Dervish}{درویش} moves between poverty, patronage, and ethical self-positioning; \lex{shah}{King}{شاه} moves between epic sovereignty, panegyric address, and moral kingship. These are not random alternations. They show how poetic voice redistributes a shared vocabulary across different social and symbolic economies.

Table~\ref{tab:focus-comparison} sharpens the internal contrast among the five recurrent reference words. \lex{del}{Heart}{دل} and \lex{shab}{Night}{شب} sit above the century baseline as hubs more consistently than their raw drift values alone would suggest, which helps explain why both remain so legible despite ongoing rewiring. \lex{khaak}{Earth}{خاک} stays closer to the period mean and changes more with poet-specific ethical stance: humility, abasement, threshold imagery, and materiality do not disappear, but different poets weight them very differently. \lex{mey}{Wine}{می} combines high drift and high turnover with a weaker sustained hub deviation, which is exactly what one expects from a term dispersed across convivial, mystical, satirical, and formulaic lyric registers. \lex{baadeh}{Wine}{باده}, by contrast, keeps a narrower convivial organization, so its semantic history remains easier to narrate even when it changes.

\begin{table}[!htbp]
    \centering
    \small
    \caption{Comparative profile of the five recurrent reference words. Mean hub reports raw degree centrality across viable century graphs; hub deviation reports the average departure from the century mean, so positive values indicate words that tend to sit above the prevailing hub level of their period.}
    \label{tab:focus-comparison}
    \begin{tabularx}{\textwidth}{@{}l c c c c l >{\raggedright\arraybackslash}X@{}}
\toprule
Word & Drift & Turnover & Mean hub & Hub dev. & Pressure & Literary mechanism \\
\midrule
\latinlex{khaak}{Earth} & 0.83 & 0.80 & 0.002 & -0.000 & Mixed & Threshold, abasement, and materiality are redistributed by poet-specific ethical stance. \\
\latinlex{shab}{Night} & 0.83 & 0.69 & 0.002 & +0.000 & More time-sensitive & The nocturnal field stays shared, but its historical weighting shifts with separation, darkness, and vigil. \\
\latinlex{mey}{Wine} & 0.82 & 0.94 & 0.001 & -0.001 & More time-sensitive & A broad symbolic probe absorbs convivial, mystical, satirical, and lyrical registers at once. \\
\latinlex{baadeh}{Wine} & 0.90 & 0.78 & 0.002 & +0.001 & More time-sensitive & A narrower wine lexicon keeps a more coherent convivial cluster across the corpus. \\
\latinlex{del}{Heart} & 0.74 & 0.80 & 0.002 & +0.000 & More time-sensitive & An enduring affective center persists while its internal attachments are repeatedly rewired. \\
\bottomrule
\end{tabularx}

\end{table}

The same contrast clarifies the affective and mystical layers. \lex{del}{Heart}{دل}, \lex{jaan}{Soul}{جان}, and \lex{gham}{Sorrow}{غم} preserve an inward cluster of hurt, endurance, consolation, and self-offering. \lex{eshgh}{Love}{عشق} is less compact because love traverses lyric, didactic, and mystical uses more freely. The explicitly mystical words are narrower still in distribution. \lex{fanaa}{Annihilation}{فنا}, \lex{baqaa}{Subsistence}{بقا}, and \lex{haqiqat}{Truth}{حقیقت} often move together, but they become most active where Sufi discourse itself is foregrounded. \lex{tariqat}{Path}{طریقت} and \lex{soofi}{Sufi}{صوفی} intensify that pattern: when they appear, they enter strongly doctrinal or polemical neighborhoods rather than the broad symbolic field occupied by words such as \lex{shab}{Night}{شب} or \lex{eshgh}{Love}{عشق}. In literary terms, technical mystical vocabulary is less uniformly shared than emblematic mystical imagery because doctrinal vocabulary is concentrated in particular poets and argumentative settings, whereas emblematic mystical imagery travels more freely across lyric convention.

\begin{figure}[!tbp]
    \centering
    \includegraphics[width=\textwidth]{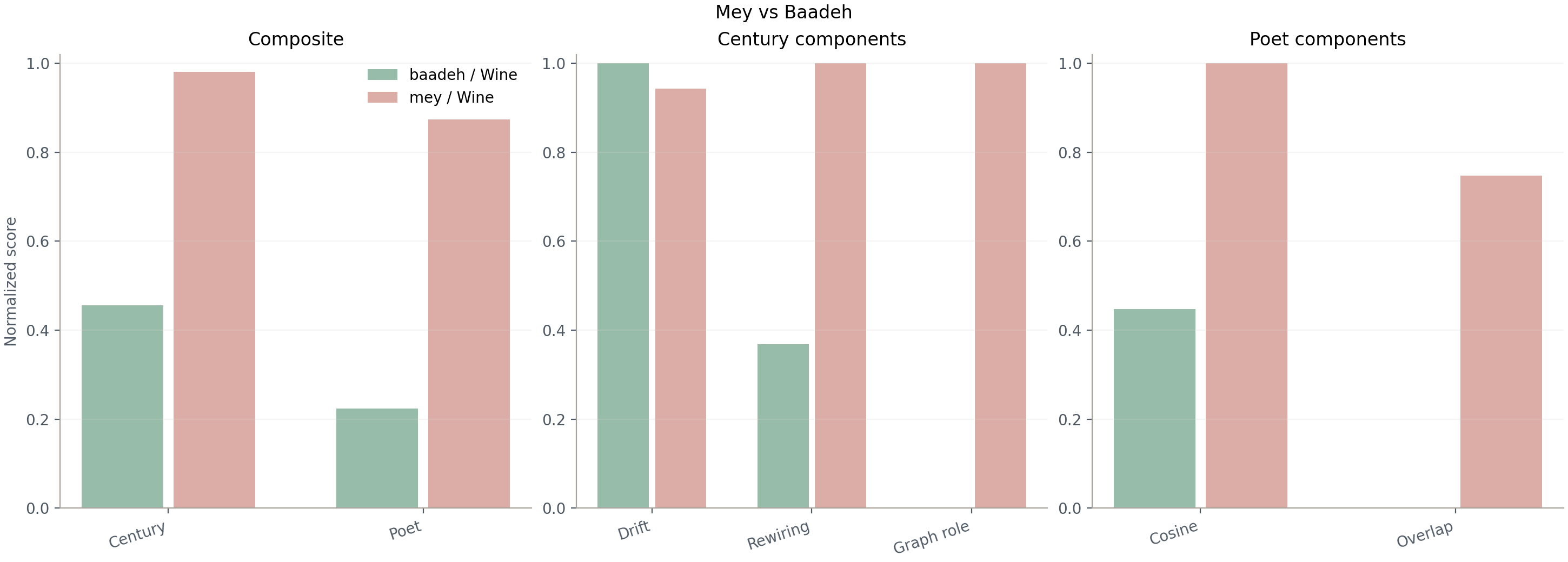}
    \caption{Direct comparison between \latinlex{mey}{Wine} and \latinlex{baadeh}{Wine}. Both remain historically active, but \latinlex{baadeh}{Wine} keeps a narrower convivial neighborhood while \latinlex{mey}{Wine} disperses across a broader symbolic field.}
    \label{fig:mey-baadeh}
\end{figure}

The wine pair brings the interpretive and methodological stakes into one frame. Figure~\ref{fig:mey-baadeh} makes clear that \lex{mey}{Wine}{می} and \lex{baadeh}{Wine}{باده} should not be treated as interchangeable probes. \lex{mey}{Wine}{می} is semantically expansive. It drifts strongly, rewires heavily, and varies substantially across poets. Its neighbors sometimes remain within familiar wine imagery, but they also spill into a broader field that becomes harder to stabilize because the term moves easily between literal wine, mystical intoxication, social satire, and formulaic lyric diction. \lex{baadeh}{Wine}{باده}, by contrast, keeps returning to a more coherent convivial cluster involving \lex{saaghi}{Cupbearer}{ساقی}, \lex{jam}{Cup}{جام}, \lex{qadah}{Cup}{قدح}, and \lex{sharaab}{Wine}{شراب}. That narrower semantic footprint does not make \lex{baadeh}{Wine}{باده} uninteresting. It makes it more interpretable. The cleaner probe is the one whose rewiring stays narratable within a wine symbolism that remains deeply embedded in Persian lyric convention \citep{SaeidiUnwin2005}.

\subsection{Time and Poet}

\begin{figure}[!tbp]
    \centering
    \begin{subfigure}[t]{\textwidth}
        \centering
        \includegraphics[width=0.96\textwidth,height=0.34\textheight,keepaspectratio]{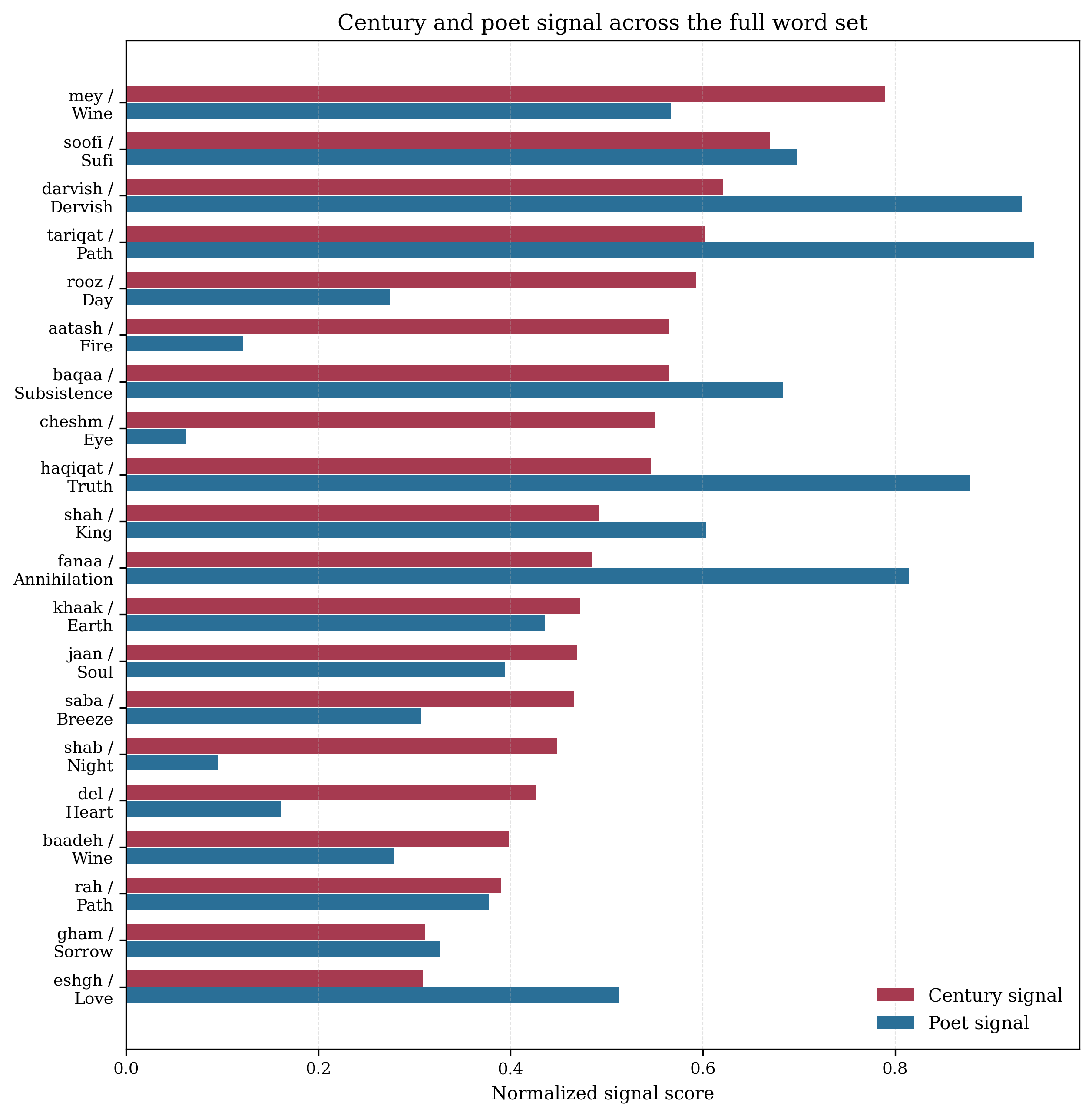}
        \caption{Composite signal scores.}
    \end{subfigure}

    \vspace{0.35em}

    \begin{subfigure}[t]{\textwidth}
        \centering
        \includegraphics[width=0.90\textwidth,height=0.34\textheight,keepaspectratio]{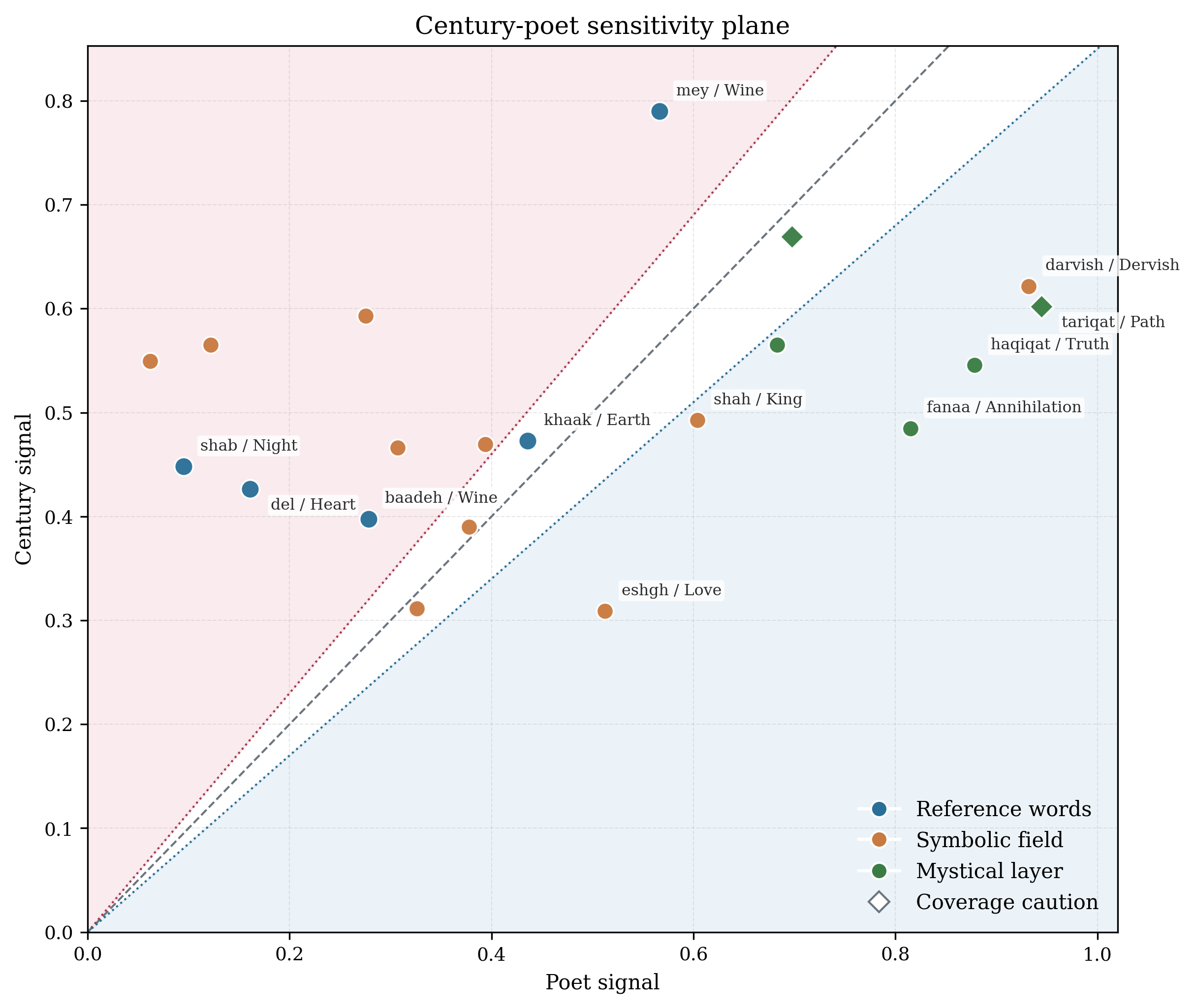}
        \caption{Sensitivity regions.}
    \end{subfigure}
    \caption{Direct comparison of century-side and poet-side signal for all twenty target words. The bar panel compares the composite signal scores, and the plane situates the lexical panel across time-sensitive, mixed, and poet-sensitive regions.}
    \label{fig:century-poet}
\end{figure}

The century-versus-poet comparison consolidates the argument at the scale of the full panel. Figure~\ref{fig:century-poet} shows that the corpus supports several distinct pressure profiles at once. Time-sensitive words include \lex{shab}{Night}{شب}, \lex{rooz}{Day}{روز}, \lex{cheshm}{Eye}{چشم}, \lex{aatash}{Fire}{آتش}, \lex{saba}{Breeze}{صبا}, \lex{baadeh}{Wine}{باده}, and \lex{jaan}{Soul}{جان}. These belong to symbolic fields that remain broadly shared across poets even while their internal emphasis changes across centuries. Poet-sensitive words include \lex{khaak}{Earth}{خاک}, \lex{eshgh}{Love}{عشق}, \lex{darvish}{Dervish}{درویش}, \lex{shah}{King}{شاه}, \lex{fanaa}{Annihilation}{فنا}, \lex{baqaa}{Subsistence}{بقا}, \lex{haqiqat}{Truth}{حقیقت}, and \lex{tariqat}{Path}{طریقت}. These words are more exposed to authorial specialization, courtly convention, or explicitly Sufi discourse.

The mixed cases matter just as much. \lex{mey}{Wine}{می}, \lex{rah}{Path}{راه}, and \lex{gham}{Sorrow}{غم} do not fall near the center because they are weakly informative. They do so because both pressures remain active at once. \lex{mey}{Wine}{می} is diffuse across centuries and poets alike because the term can be recruited into multiple registers without stabilizing in one dominant field. \lex{rah}{Path}{راه} oscillates between literal road, spiritual way, and ethical orientation, so its historical change is inseparable from shifts in genre and didactic address. \lex{gham}{Sorrow}{غم} keeps a durable grief vocabulary but shifts with lyric and moral context. The comparison therefore cautions against any binary opposition between diachrony and authorship. Persian poetic vocabulary is distributed across both, and the graph evidence shows that these pressures alter not only how far words move, but which companions they carry with them.

\subsection{Lexical Fields}

The twenty-word panel forms a single semantic ecology. Table~\ref{tab:lexical-panel} consolidates the entire set. The affective field divides between durable interior terms such as \lex{del}{Heart}{دل}, \lex{jaan}{Soul}{جان}, and \lex{gham}{Sorrow}{غم}, and more poet-shaped desire terms such as \lex{eshgh}{Love}{عشق} and \lex{cheshm}{Eye}{چشم}. The elemental and temporal words \lex{aatash}{Fire}{آتش}, \lex{saba}{Breeze}{صبا}, and \lex{rooz}{Day}{روز} remain strongly historical because they are shared symbolic resources whose rhetorical weighting changes with period. The courtly pair \lex{darvish}{Dervish}{درویش} and \lex{shah}{King}{شاه} shows the sharpest poet-side contrast: one organizes poverty and renunciation against wealth and power, the other organizes sovereignty, service, and ethical rule.

\begin{table}[!htbp]
    \centering
    \footnotesize
    \setlength{\tabcolsep}{4pt}
    \renewcommand{\arraystretch}{0.96}
    \caption{Summary of the full lexical panel. Drift and turnover are mean adjacent century-side values; the graph-role column summarizes the dominant trajectory shape, and the pressure column reports whether century-side or poet-side differentiation is stronger.}
    \label{tab:lexical-panel}
    \begin{tabularx}{\textwidth}{@{}l l c c >{\raggedright\arraybackslash}X >{\raggedright\arraybackslash}X@{}}
\toprule
Word & Field & Drift & Turnover & Graph role & Dominant pressure \\
\midrule
\multicolumn{6}{@{}l@{}}{\textit{Reference words}} \\
\latinlex{khaak}{Earth} & Reference words & 0.83 & 0.80 & Community-migrant & Mixed \\
\latinlex{shab}{Night} & Reference words & 0.83 & 0.69 & Community-migrant & More time-sensitive \\
\latinlex{mey}{Wine} & Reference words & 0.82 & 0.94 & Community-migrant & More time-sensitive \\
\latinlex{baadeh}{Wine} & Reference words & 0.90 & 0.78 & Community-migrant & More time-sensitive \\
\latinlex{del}{Heart} & Reference words & 0.74 & 0.80 & Community-migrant & More time-sensitive \\
\midrule
\multicolumn{6}{@{}l@{}}{\textit{Symbolic field}} \\
\latinlex{eshgh}{Love} & Symbolic field & 0.88 & 0.73 & Community-migrant & More poet-sensitive \\
\latinlex{jaan}{Soul} & Symbolic field & 0.81 & 0.79 & Community-migrant & More time-sensitive \\
\latinlex{cheshm}{Eye} & Symbolic field & 0.85 & 0.73 & Community-migrant & More time-sensitive \\
\latinlex{aatash}{Fire} & Symbolic field & 0.83 & 0.75 & Community-migrant & More time-sensitive \\
\latinlex{darvish}{Dervish} & Symbolic field & 0.91 & 0.93 & Community-migrant & More poet-sensitive \\
\latinlex{shah}{King} & Symbolic field & 0.87 & 0.90 & Community-migrant & More poet-sensitive \\
\latinlex{rah}{Path} & Symbolic field & 0.89 & 0.80 & Community-migrant & Mixed \\
\latinlex{saba}{Breeze} & Symbolic field & 0.92 & 0.80 & Community-migrant & More time-sensitive \\
\latinlex{gham}{Sorrow} & Symbolic field & 0.82 & 0.75 & Community-migrant & Mixed \\
\latinlex{rooz}{Day} & Symbolic field & 0.84 & 0.80 & Community-migrant & More time-sensitive \\
\midrule
\multicolumn{6}{@{}l@{}}{\textit{Mystical layer}} \\
\latinlex{fanaa}{Annihilation} & Mystical layer & 0.89 & 0.86 & Community-migrant & More poet-sensitive \\
\latinlex{baqaa}{Subsistence} & Mystical layer & 0.92 & 0.88 & Community-migrant & More poet-sensitive \\
\latinlex{tariqat}{Path} & Mystical layer & 0.89 & 0.95 & Low-data caution & More poet-sensitive (caution) \\
\latinlex{haqiqat}{Truth} & Mystical layer & 0.87 & 0.94 & Community-migrant & More poet-sensitive \\
\latinlex{soofi}{Sufi} & Mystical layer & 0.93 & 0.95 & Low-data caution & Mixed (caution) \\
\bottomrule
\end{tabularx}

\end{table}

The mystical layer is especially instructive because it is internally stratified. \lex{fanaa}{Annihilation}{فنا}, \lex{baqaa}{Subsistence}{بقا}, and \lex{haqiqat}{Truth}{حقیقت} often cluster together with being, non-being, gnosis, and perception, yet they remain more poet-sensitive than the core temporal words because technical Sufi discourse is concentrated in specific authors and milieux \citep{Chittick2017Rumi,Schimmel1975}. \lex{tariqat}{Path}{طریقت} and \lex{soofi}{Sufi}{صوفی} narrow the distribution further. When they appear, they do so in strongly doctrinal or polemical neighborhoods. Persian literary tradition thus preserves mystical vocabulary in more than one form: some terms circulate widely as symbolic concepts, while others remain tied to explicitly Sufi diction.

\subsection{Human Validation of Semantic Change}

To check whether the detected changes remained legible from within Persian literary usage, the study also included a modest human validation exercise on 100 sampled word instances distributed across centuries and poets. A domain expert in Persian literature assessed whether the model-indicated shifts aligned with plausible literary interpretation. The exercise was qualitative rather than fully annotational, so the article reports broad agreement rather than a single headline coefficient.

The expert review largely confirmed the patterns emphasized by the computational analysis. For \lex{khaak}{Earth}{خاک}, sampled contexts supported movement between abasement, thresholded nearness, dust imagery, and more expansive material or cosmological uses. For \lex{shab}{Night}{شب}, the review confirmed that the nocturnal field persists even as the balance among darkness, temporal opposition, and separation shifts historically. For \lex{del}{Heart}{دل}, \lex{jaan}{Soul}{جان}, and \lex{gham}{Sorrow}{غم}, the expert repeatedly noted continuity of affective force alongside changes in the pressure placed on bodily, ethical, and intimate associations. Samples of \lex{eshgh}{Love}{عشق}, \lex{darvish}{Dervish}{درویش}, and \lex{shah}{King}{شاه} likewise supported the poet-side distinctions identified above, while \lex{fanaa}{Annihilation}{فنا}, \lex{baqaa}{Subsistence}{بقا}, and \lex{haqiqat}{Truth}{حقیقت} were repeatedly judged most plausible in explicitly Sufi argumentative settings. The wine pair remained especially clarifying: \lex{baadeh}{Wine}{باده} was consistently judged the narrower convivial term, while \lex{mey}{Wine}{می} appeared across a looser and more variable figurative range.

This validation should not be overstated. It does not convert the study into a supervised benchmark, nor does it eliminate interpretive ambiguity. What it does show is that the main computational signals remain recognizable to expert literary judgment. That matters because the article's claim is not only that the models detect change, but that the detected changes correspond to plausible patterns of poetic usage.

\section{Discussion}

The results support a relational theory of lexical history. In Persian poetry, words do not simply move from one meaning to another in a linear sequence. They persist through recurrent literary traditions while changing the neighborhoods that make them legible. A word may keep its thematic center and still alter the work it performs because its nearest companions, bridge ties, and community memberships have changed. That is why graph-based rewiring provides a better interpretive vocabulary than drift alone. Drift can show that a word moved; it cannot by itself show whether the movement reflects a field-wide historical reorganization, a slice-bound fluctuation, or the relative differentiation of one word from its contemporaries. The raw and century-centered graph views separate the first and third mechanisms, while the global reference baseline helps isolate the second.

The twenty-word panel sharpens that claim by showing that different lexical domains reassemble in different ways. Affective vocabulary such as \lex{del}{Heart}{دل}, \lex{jaan}{Soul}{جان}, and \lex{gham}{Sorrow}{غم} remains durable because interior suffering, desire, and consolation are deeply shared resources in Persian lyric. Yet even here continuity is not immobility: those words keep changing the balance between wound, patience, nearness, and self-offering. Their raw centrality often rises together with the broader field, but the centered view shows that each term occupies a different relative position within that collective rise. Elemental and temporal words such as \lex{aatash}{Fire}{آتش}, \lex{rooz}{Day}{روز}, and \lex{saba}{Breeze}{صبا} are more historical because their symbolic work is repeatedly redistributed by period. Courtly and ascetic vocabulary such as \lex{shah}{King}{شاه} and \lex{darvish}{Dervish}{درویش} is more poet-shaped because each poet bends those terms toward a different social imagination, from ethical kingship and panegyric praise to renunciation, marginality, and spiritual poverty.

The mystical layer clarifies the same point from another angle. \lex{fanaa}{Annihilation}{فنا}, \lex{baqaa}{Subsistence}{بقا}, and \lex{haqiqat}{Truth}{حقیقت} travel together conceptually, but they do not circulate evenly across the corpus. Their strongest signatures appear where Sufi discourse becomes explicit, which is why they look more poet-sensitive than broadly shared temporal or elemental words. \lex{tariqat}{Path}{طریقت} and \lex{soofi}{Sufi}{صوفی} narrow the distribution further. The result is a distinction between symbolically pervasive mystical imagery and technically inflected mystical vocabulary. Persian poetry preserves both, but they leave different statistical traces because they occupy different literary functions: emblematic mystical language can travel across lyric convention, whereas doctrinal vocabulary remains concentrated in particular milieux, pedagogies, and arguments.

The contrast between \lex{mey}{Wine}{می} and \lex{baadeh}{Wine}{باده} adds a final methodological lesson. A culturally central word is not always the best lexical probe. Broad probes absorb multiple registers, formulae, and ambiguities at once. Narrower probes can yield a clearer history without being literarily trivial. Here \lex{baadeh}{Wine}{باده} remains more legible than \lex{mey}{Wine}{می} precisely because its rewiring stays more disciplined. The same principle extends beyond wine vocabulary: the history of a word becomes interpretable when its changing neighbors still form a narratable field, and graph structure is what makes that narratability visible.

\section{Limitations}

Several limitations shape the scope of the findings. The first is temporal unevenness. Century 3 remains an unstable slice. The article keeps the early century visible because historical depth matters, but the strongest century-level claims depend more heavily on the denser middle and later slices. The global-reference comparison improves interpretability by distinguishing slice-bound fluctuation from departures that also remain visible against the full-corpus baseline, yet it cannot recreate lexical evidence that the sparse slice never contained. The caution therefore remains substantive rather than merely procedural.

A second limitation concerns lexical ambiguity. \lex{mey}{Wine}{می} is revealing precisely because it is broad, but that breadth also makes it noisier than the other probes. Some of its instability reflects genuine semantic richness; some of it reflects the difficulty of forcing a highly polysemous literary term into a single probe role. The contrast with \lex{baadeh}{Wine}{باده} makes this limitation visible rather than hidden, yet the problem remains real.

The third limitation is representational. Static embeddings are well suited to broad comparative structure, but they smooth over local contextual nuance. Poetry often depends on precisely the kinds of syntagmatic and rhetorical shifts that static models compress. A contextual model might distinguish registers and figurative turns that remain blended here. Likewise, the selected poet set is intentionally limited. It supports disciplined comparison, but it cannot stand in for the whole history of Persian poetic authorship.

Finally, lexical restriction is itself an intervention. By concentrating on the strongest recurring neighbors, the analysis gives less weight to weaker or more episodic ties that may still matter in particular poems. Especially sparse or highly polysemous terms should therefore be interpreted with restraint.

\section{Conclusion}

This study has argued that lexical meaning in Persian poetry changes not only through movement in embedding space but through the rewiring of semantic neighborhoods. The full twenty-word panel shows that this rewiring takes different forms in different lexical domains. \lex{shab}{Night}{شب}, \lex{rooz}{Day}{روز}, \lex{aatash}{Fire}{آتش}, and \lex{saba}{Breeze}{صبا} are more strongly historical. \lex{khaak}{Earth}{خاک}, \lex{eshgh}{Love}{عشق}, \lex{darvish}{Dervish}{درویش}, \lex{shah}{King}{شاه}, and several explicitly mystical terms are more strongly shaped by poetic voice. \lex{del}{Heart}{دل}, \lex{gham}{Sorrow}{غم}, \lex{rah}{Path}{راه}, and \lex{mey}{Wine}{می} sit between those poles, but for different reasons: some preserve durable semantic centers, while others remain unstable on both axes.

For Digital Humanities, the broader contribution lies in the form of explanation rather than in any single metric. Graph-based semantic analysis makes it possible to describe lexical history in terms closer to literary interpretation: persistence, migration, mediation, and local reshaping. It shows how computational models can support humanistic argument when they stay close to the structure of the lexical field rather than flattening it into a single distance score. What emerges most clearly from the twenty-word panel is that meaning in Persian poetry does not survive by standing still. It survives by being reassembled.

Data, trained models, and analysis code will be made available through the project repository.

\section*{Acknowledgements}

This study owes a quiet debt to the custodians of Persian literary memory: the editors, scholars, and archivists whose labors have carried these poems across manuscripts, printed editions, and digital collections. We are equally grateful for the conversations, formal and informal, that continue to keep Persian poetry intellectually alive, where philological care and literary imagination sharpen one another. Any attempt to follow a word across centuries begins from that inherited generosity.

\begingroup
\sloppy
\printbibliography[title={References}]
\endgroup

\end{document}